\documentclass[conference]{IEEEtran}
\IEEEoverridecommandlockouts

\def\BibTeX{{\rm B\kern-.05em{\sc i\kern-.025em b}\kern-.08em
    T\kern-.1667em\lower.7ex\hbox{E}\kern-.125emX}}
\usepackage{cite}

\usepackage[a4paper,
            left=1in,
            right=1in,
            top=1in,
            bottom=2.6cm
            ]{geometry}

\usepackage{hyperref}
\usepackage[pdftex]{graphicx}
\usepackage{multirow}
\usepackage{algorithm,algpseudocode,float}
\usepackage{booktabs}
\usepackage{hhline}

\usepackage{cite}
\usepackage{amsmath,amssymb,amsfonts}
\usepackage{graphicx}
\usepackage{textcomp}
\usepackage{xcolor}
\usepackage{url}
\usepackage{subfig} % for multiple image

\usepackage{array}
\begin{document}
\title{Representation Learning on Out of Distribution in Tabular Data\\
\thanks{
\textcopyright~\the\year~IEEE. Personal use of this material is permitted. 
Permission from IEEE must be obtained for all other uses, in any current or future media, including reprinting/republishing this material for advertising or promotional purposes, creating new collective works, for resale or redistribution to servers or lists, or reuse of any copyrighted component of this work in other works.
}
}

\author{\IEEEauthorblockN{1\textsuperscript{st} Achmad Ginanjar}
\IEEEauthorblockA{\textit{School of EECS} \\    
\textit{The University of Queensland}\\
Brisbane, Queensland, Australia \\
a.ginanjar@uq.edu.au}
\and
\IEEEauthorblockN{2\textsuperscript{nd} Xue Li}
\IEEEauthorblockA{\textit{School of EECS} \\
\textit{The University of Queensland}\\
Brisbane, Queensland, Australia}
\and
\IEEEauthorblockN{3\textsuperscript{rd} Priyanka Singh}
\IEEEauthorblockA{\textit{School of EECS} \\
\textit{The University of Queensland}\\
Brisbane, Queensland, Australia}
\and
\IEEEauthorblockN{4\textsuperscript{th} Wen Hua}
\IEEEauthorblockA{\textit{Department of Computing} \\
\textit{The Hong Kong Polytechnic University}\\
Hong Kong}
}

\maketitle
% \begin{flushleft}
% \scriptsize
% \textcopyright~\the\year~IEEE. Personal use of this material is permitted. 
% Permission from IEEE must be obtained for all other uses, in any current or future media, including reprinting/republishing this material for advertising or promotional purposes, creating new collective works, for resale or redistribution to servers or 
% lists, or reuse of any copyrighted component of this work in other works.
% \end{flushleft}

\begin{abstract}

The open world assumption in model development indicates that a model may not have enough information to effectively manage data that is completely different or out of distribution (OOD). When a model encounters OOD data, its performance can deteriorate significantly. To enhance a model's ability to handle OOD data, generalisation techniques can be employed, such as adding noise, which is easily implemented using deep learning methods. However, many advanced machine learning models are resource-intensive and optimised for use with specialised hardware like GPU, which may not always be accessible to users with limited hardware capabilities. To provide a deeper understanding and practical solutions for handling OOD data, this study explores detection, evaluation, and prediction tasks within the context of OOD on tabular datasets, specifically using common consumer hardware (CPUs). The study demonstrates how users can identify OOD data from available datasets and offers guidance on evaluating the selection of OOD data through straightforward experiments and visualisations. Additionally, this study introduces a technique called Tabular Contrast Learning (TCL), a representation learning technique which is specifically designed for tabular prediction tasks. TCL achieves better results compared to heavier models while being more efficient, even when trained on non-specialised hardware. This makes it particularly beneficial for general machine learning users who face computational constraints. The results show that TCL outperforms other contrastive learning methods and various deep learning models in classification tasks.
\end{abstract}

\section{Introduction}
The concept of open-world assumption in model development means that a model may not have enough information to effectively handle data that is completely different or out of distribution (OOD). When a model meets OOD data, it may suffer a significant decrease in performance \cite{Hsu2020,OODBaseline}. To handle this, model generalisation by introducing noise can be used, which can be achieved easily with deep learning. However, advanced deep learning algorithms such as FT-Transformer benefit from the advancement of specialised hardware such as GPU or TPU \cite{Hwang2018Background}, this type of hardware is not always available to the general user \cite{ahmed2020dedemocratizationaideeplearning}.  These demand the user to find the best way to deal with these challenges and emphasise the importance of our research.

While out-of-distribution (OOD) detection has been extensively studied \cite{Lee2020}, the challenge of prediction tasks for OOD data, particularly in tabular datasets, remains under explored. Significant progress has been made in OOD detection with algorithms like MCCD \cite{multiClass}, OpenMax \cite{Bendale2016}, Monte Carlo Dropout \cite{Gal2016}, and ODIN \cite{Liang2017}. However, the study of prediction tasks on OOD for tabular data is limited. Tree-based classical models are known to be reliable for tabular data  \cite{Grinsztajn2022}, but our experiments show that these models exhibit a decrease in performance when dealing with OOD data.

In this study, we made several contributions. First, we show step by step how to implement existing methods for detecting, separating, evaluating, and visualising out-of-distribution (OOD) data using real-world datasets. Second, we assess the performance of existing tabular machine learning algorithms in handling OOD data. Lastly, we introduce a new approach called TCL, which provides efficiency and flexibility while achieving comparable performance.

\begin{figure*}[]
    \centering
    \includegraphics[width=0.8\linewidth]{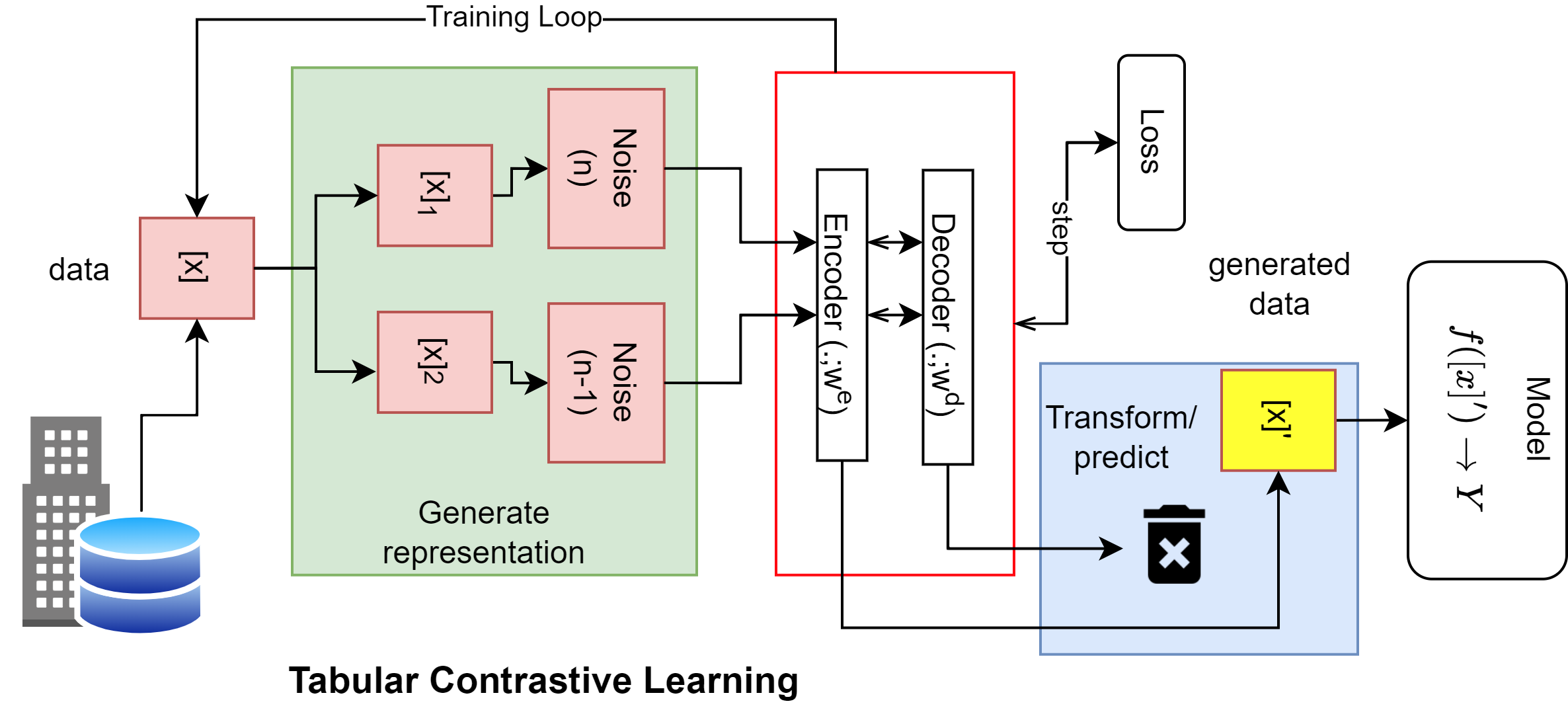}
    \caption{Tabular Contrastive Learning (TCL). The data $[x]$ is duplicated $([x]_1,[x]_2)$, and noise is added. Both duplicates are then encoded and decoded to compute the loss. During inference, TCL only uses the encoder to produce new data $[x]'$ that enhances supervised learning performance $f([x]') \rightarrow Y $. The decoder is omitted during inference and is used only for training.}
    \label{fig:tcl}
\end{figure*}

Tabular Contrast Learning (TCL), Figure \ref{fig:tcl}, is a local adaptation of Contrastive Federated Learning (CFL) \cite{anonymizedMethod} designed for prediction tasks on tabular datasets for a general user. TCL is based on the principles of contrastive learning \cite{Chen2020SimClr,subtab} but is optimised for tabular data structures. TCL approach offers several advantages, e.g. 
\textbf{Efficiency}: TCL is designed to be faster and more compact compared to current state-of-the-art models,
\textbf{Flexibility}: TCL can be integrated with various supervised learning algorithms and 
\textbf{Performance}: TCL achieves competitive performance.
 
Our experiment demonstrates that TCL delivers performance and efficiency \cite{huang2017speedaccuracy} (defined by a higher speed/accuracy trade-off score)  compared to other models. 

\section{Related Work}
\subsection{OOD detection}
\textbf{OpenMax} \cite{Bendale2016} uses the concept of Meta-Recognition to estimate the probability that an input belongs to an unknown class. OpenMax characterises the failure of the recognition system and handles unknown/unseen classes during operation. In deep learning, SoftMax calculates  as 
\begin{gather}
    \begin{split}
        P(y=j|x) = \dfrac{e^{v_j(x)}}{\sum^N_{i=1} e^{v_i(x)} }
    \end{split}
\end{gather}

OpenMax recognises that in out-of-distribution (OOD), the denominator of the SoftMax layer does not require the probabilities to sum to 1.

\textbf{TemperatureScaling} \cite{platt1999probabilistic} is a single-parameter variant of the Platt scaling. In a study by Guo\textit{ et al.}  \cite{Guo2017}, despite its simplicity, temperature scaling is effective in calibrating a model for deep learning. This also suggests that temperature scaling can be used to detect OOD.  Our study uses these two approaches to separate OOD from the dataset and use it as validation data.

\textbf{Multi-class classification, deep neural networks, Gaussian discriminant analysis (MCCD)} \cite{multiClass} is OOD detection algorithm based deep neural network that claim to have better classification inference performance. It focuses on finding spherical decisions across classes. 

Our work mainly uses OpenMax and TemperatureScaling. While the original algorithms are not new, both algorithms have the latest update and better support under pytorch-ood \cite{kirchheim2022pytorch} compared to MCCD \cite{deepMCCD}.

\subsection{Tabular data prediction}
    
\textbf{Neural Network-based Methods:}
    Multilayer Perceptron (MLP) \cite{mlp,tabr}: A straightforward deep learning approach for tabular data.
    Self-Normalising Neural Networks (SNN).  \cite{snn}: Uses SELU activation to train deeper networks more effectively.

\textbf{Advanced Architectures:}
    Feature Tokeniser Transformer / FT-Transformer \cite{vaswani2017attentiontransformer}: Adapts the transformer architecture for tabular data, consistently achieving high performance.
    Residual Network / ResNet \cite{restnet}: Utilises parallel hidden layers to capture complex feature interactions.
    Deep \& Cross Network  / DCN V2 \cite{DCNV2}: Incorporates a feature-crossing module with linear layers and multiplications.
    Automatic Feature Interaction / AutoInt \cite{AutoInt}: Employs attention mechanisms on feature embeddings.
    Neural Oblivious Decision Ensembles / NODE \cite{NODE}: A differentiable ensemble of oblivious decision trees.
    Tabular Network / TabNet \cite{TabNet}: Uses a recurrent architecture with periodic feature weight adjustments. Focuses on the attention framework.

\textbf{Ensemble Methods:}
    GrowNet \cite{GrowNet}: Applies gradient boosting to less robust MLPs, primarily for classification and regression tasks.

\textbf{Gradient Boosting Decision Tree (GBDT) } \cite{Grinsztajn2022} :
    XGBoost :A tree-based ensemble method that uses second-order gradients and regularisation to prevent overfitting while maximising computational efficiency.
    LightGBM :A fast and memory-efficient boosting framework that uses histogram-based algorithms and leaf-wise tree growth strategy for faster training.
    CatBoost :A gradient boosting implementation specifically optimised for categorical features with built-in ordered boosting to reduce prediction shift.

These models have shown varying degrees of success in tabular data prediction. However, their performance on OOD data remains a critical area for investigation. We include the mentioned models as our base models.

\subsection{Tabular contrastive learning}
    \textbf{SubTab} \cite{subtab} and \textbf{SCARF} \cite{scarf} are a contrastive learning model tailored for tabular datasets. Similar to the fundamental concept of contrastive learning for the image of SimCLR \cite{Chen2020SimClr}. SubTab and SCARF calculate contrastive loss using cosine or euclidean distance.
    \textbf{CFL} \cite{anonymizedMethod} is a federated learning algorithm proposed to tackle vertical partition within data silos. CFL explores the possibility of implementing contrastive learning within vertically partitioned data without the need for data sharing. CFL merge the weight by understanding that the data came from a global imaginary dataset, which is vertically partitioned. CFL uses contrastive learning as a medium for black box learning. CFL focuses on collaborative learning across silos. In this study, we study learning from local data with OOD, a problem that is yet to be explored by CFL. CFL focuses on a Federated learning network, while ours is common tabular data.  CFL, while exhibiting a similar name, uses partial data augmentation as part of the federated learning concept and is similar to image contrastive learning. TCL, in the other hand use full matrix augmentation to support tabular data.

\section{Problem Formulation}
\subsection{Definition}
\textbf{Definition 1: Tabular Data.}
Let $D \in R^{n×d}$ be a tabular dataset with $n$ samples and $d$ features, and $Y \in R^n$ be the corresponding labels. Tabular data is characterised by its structured format, where each row represents a sample and each column represents a feature.  

\textbf{Definition 2: Out-of-Distribution Prediction.}
Given a model trained on in-distribution data:
\begin{gather}
    \begin{split}
        D_{in}=\{(x_i,y_i)|x_i \in X_{in},y_i \in Y_{in}\}
    \end{split}
\end{gather}
 Where $X_{in}$ follows a distribution $p_{in}$, the task is to make accurate predictions on OOD data $X_{ood}$ that follows a different distribution $p_{ood}$, where $p_{ood} \neq p_{in}$.

\textbf{Definition 3: Efficiency-Accuracy Trade-off.}
We use a speed/accuracy trade-off \cite{huang2017speedaccuracy} from the total performance matrix. Let $T$ be the total performance , then $ T=\frac{P}{t}$ for classification or  $T=\frac{1/P}{t} $ for regression.

A performance evaluation $P$ used is F1 or RMSE for the prediction task, and the time $t$ in seconds for the duration of training. The $P$ is used to obtain the standard performance of a model. The $t$ is used to evaluate the time it takes to train a model. A smaller $t$ means a smaller resource to find the best model by tuning the hyperparameters.  The adjustment $\frac{1}{P}$ for regression is necessary because in regression tasks, RMSE is used as the performance metric, and smaller values are better. RMSE was selected because it provide positive value which is simple and robust explanation of error.

\subsection{Problem Statement}
In machine learning, the goal is to build a model $f:x\rightarrow y$  that generalises unseen data. However, if the model is exposed to samples outside the distribution during inference, it may make unreliable predictions or exhibit unexpected behaviour.

Formally, a prediction task can be defined as finding a model $f:(.)$ that minimises the expected error over a dataset   $D_{in}$ with a distribution $p_{in}$.  This can be defined as 
\begin{gather}
    \begin{split}
        \min_f \text{Error}(x,y) _{D} = [L(f(x), y)]
    \end{split}
    \label{math:error}
\end{gather}
 Where $[L(\cdot)]$ is the expected loss value that measures the distance between the prediction of the model $f(x)$ and the true label $y$. Formula \ref{math:error} subject to:

\begin{enumerate}
    \item $t_{train} \leq t_{\max}$ (training time constraint)
    \item $M_{usage} \leq M_{available}$ (memory constraint)
    \item $P_{model} \leq P_{max}$ (parameter constraint)
    \item $t_{inference} \leq t_{max}$ (inference time constraint)
    \item $|P(f(x_{D_{in}}))- P(f(x_{D_{ood}}))|\leq \delta$ (OOD performance degradation constraint)
\end{enumerate}

A bigger $E$ means poor model performance $P$ or can be denoted as $E\uparrow = P\downarrow$. When a different distribution $p_{ood}$ is introduced to a model, the performance decreases.
\begin{gather}
    \begin{split}
        P(f((x)_{D_{in}})) > P(f((x)_{D_{ood}}))
    \end{split}
\end{gather} 

The time $t$ to find the best model should also be considered. A time-consuming model takes more resources to train and tune. When dealing with a large dataset, the time used to train and tune a model is a big concern. A larger and longer model does not always equal better performance $P$. This can be written as $t\uparrow \ne P\uparrow$.

We evaluated the total/overall performance of the models when dealing with tabular dataset with OOD. The overall objection can be written as 
\begin{equation}
    \begin{split}
        T &=\frac{\text{\textbf{max} }P_{ood}}{\text{\textbf{min }} t} \\
        T &= \frac{ \frac{1}{\textbf{min } E(x,y) _{D_{ood}}} }{\text{\textbf{min }} t}
    \end{split}
\end{equation}
\section{Proposed Method}
We introduce Tabular Contrastive Learning (TCL), an improved approach designed to enhance prediction tasks on tabular data, particularly with hardware limitations. 
\subsection{TCL Main Architecture}
 \textbf{Augmented Data}
 
As a contrastive learning based algorithm, TCL works based on augmented data. TCL creates two data augmentations. Denotated as $\{x^1,x^2\} = \text{Aug}(x)$.  Noise was added to these augmented data.

 \textbf{Matrix Augmentation}\

In TCL, all original data (without slicing or splitting) is utilised as a representation. Unlike previous approaches such as simCLR \cite{Chen2020SimClr} and SubTab \cite{subtab} that use slice, TCL utilises the entire original data matrix for representation. This allows for capturing more comprehensive feature interactions in tabular data. 

\textbf{Encoder-Decoder Structure}

The contrastive learning architecture includes two main components: an encoder and a decoder. The encoder transforms input data into a compressed representation called the latent space, denoted $E: x;\omega^e \rightarrow x^e$ where $\omega$ is the parameter and $e$ is the encoder notation, while the decoder reconstructs the original data from this representation, denoted $P:x^e;\omega^p \rightarrow x^p$ where $p$ is notation for the decoder. During training, both components are used, but only the encoder is employed during inference, allowing efficient compression of new data into its learned latent representation without reconstruction.

\textbf{Modified Contrastive Loss}

Loss is calculated based on augmented data, not original data. TCL  simplifies contrastive loss calculation to enhance both performance and training speed. In contrastive learning, the loss is computed based on the similarity or dissimilarity of the augmented noisy data. Since TCL deals with row-based tabular data and it is an unsupervised learning, the method used is similarity.  TCL aims to pull the noisy data points that originated from the same data. This is achieved by minimising the total loss $L_c$ between representations. During training, the total loss is calculated as follows:
 \begin{gather}
 L_t(x) = 
 (L_r(x) + L_c(x) + L_d(x)) 
\end{gather}
Where $L_r$ is the reconstruction loss, $L_c$ is the contrastive loss, and $L_d$ is the distance loss. The objective of contrastive learning is to minimise the total loss $L_t$.  When $D$ is the dataset, and sliced data $\mathcal{B} \in D$ , then:
\begin{equation}
\begin{split}
    \text{min } L_t(.;\omega^e,\omega^p) &= 
    \text{min }  \frac{1}{J} \Sigma_{j=1} ^{J} L_t( P(E(.;\omega^e);\omega^p)) \\
     % L_t(.;\omega^e,\omega^p) &= \frac{1}{J} \Sigma_{j=1} ^{J} L_t( x)
\end{split}
\end{equation}
With $w$ is weight, $P(.)$ is decoder function, and $ E(.)$ is encoder function. When $MSE(.)$ is the mean square error function, and $[x]$ is noisy data of $\mathcal{B}$,  then:
\begin{equation}
    \begin{split}
        L_r(x) &= \dfrac{1}{N} \sum_n^N MSE(\hat{x},x) \\
        L_d(x) &= \dfrac{1}{N}  \sum_n^N MSE(x^{e1},x^{e2})
    \end{split}
\end{equation}
While $L_r$ is calculated from decoded data and original data,  $L_d$ is calculated from encoded data only. We simplified the contrastive loss $L_c$  by using only the result of a dot product compared with other contrastive learning that used Euclidean distance.  
\begin{gather}
% L_c (x)  =  \dfrac{1 }{N} \sum_n^N  (−log \frac{\exp ( MSE([0],dot( x^{e1} \cdot x^{e2})) \text{/} \mathcal{T})}{\sum^K_{k=1} \exp ( MSE([0],dot(x^{e1} \cdot x^{e2} ))\text{/} \mathcal{T})})
MSE([0],dot( x^{e1} \cdot x^{e2})) \text{/} \mathcal{T}
\end{gather}

\section{TCL Algorithm}
The TCL process involves several steps. First, a mini batch of N samples is sampled from the dataset. Then, for each sample in the batch, two augmented views are created and added with noise. Although they came from the same data, both augmented data are different due to previous treatments. These augmented views are passed through an encoder network to obtain encoded representations. A decoder is then applied to the encoded representations. The loss function then calculates the difference between two augmented data. By minimising this loss, noisy data is pulled together. Because TCL applies full matrix representation, the process pulls noisy data row by row together. This results in generalised data for better inference. 

\section{Experiment}

\subsection{Datasets}
We utilised 10 diverse tabular datasets. The datasets are Adult \cite{adult_2}, Helena \cite{helenaJannis}, Jannis \cite{helenaJannis}, Higgs Small \cite{higgssmall}, Aloi \cite{aloi}, 
% Epsilon \cite{epsilon2008},
 Cover Type \cite{covtype}, California Housing \cite{ca}, Year \cite{year}, Yahoo \cite{chapelle2011yahoo}, and Microsoft \cite{microsoft}. 

\subsection{OOD Detection}
We have implemented two Out-of-Distribution (OOD) detection methods, namely OpenMax \cite{Bendale2016} and TemperatureScaling \cite{platt1999probabilistic}.  We applied OOD detection methods to each dataset to transform the data and establish thresholds. The thresholds were manually assigned by observing the graphs produced with the OOD detection algorithm. The manual assignment was done by selecting a single point on a tail of the observation. We then separated the OOD data based on the thresholds to generate two sets, $D_{in}^M$ and $D_{ood}^N$, where $M$ and $N$ are total sample in each set, $(D_{in} + D_{ood} = D)$. We expect to find $M>N$. The OOD separation is validated using linear regression. Finally, we compared the model performance with and without OOD separation, expecting to find that the performance decreased in this step $f:D_{in} > f:D_{ood}$.

The results of the experiment is two set of dataset $D_{in},D_{ood}$. While $D_{in}$ is used for train dataset, $D_{ood}$ is used for test dataset.

\subsection{Prediction ON OOD Dataset}
We experimented with 12 based models. For deep learning tabular models we use  FT-T, DCN2, GrowNet, ResNet, MLP,  AutoInt and TabR-MLP \cite{tabr}. In addition, we experimented with the recent implementation of contrastive learning for tabular data SubTab \cite{subtab} and SCARF \cite{scarf}, which comes from a similar domain to our TCL. We also did not apply FT-Transformer to some datasets. The FT-Transformer is heavy and has reached our hardware limitation. Finally, we compared our TCL with GDBT models.

\subsection{Hardware}
Our experiment used an NVIDIA H100 GPU for all models except TCL and the GBDT-based model. TCL and GBDT were trained on a CPU (Apple / AMD) to emphasise our advancement within limited hardware. The complete setting is available from the Table \ref{tab:ex_setting}.

\begin{table}[h!]
    \centering
     \caption{Environment setting for experiments. On Linux, RAM and VRAM are allocated memory. }
     \begin{tabular}{|l|l|l|l|l|} \hline 
         OS&  CPU& RAM &GPU &VRAM\\ \hline 
         Linux&  AMD EPYC&  16&H100  &16\\ \hline 
 MacOs& M4 Pro& 64& Apple GPU&shared\\ \hline
    \end{tabular}
   
    \label{tab:ex_setting}
\end{table}

\section{Result and Evaluation}

\subsection{OOD Detection}
\begin{table*}[]
% \small
\centering
\caption{The OOD detection settings.  Performances are results of models trained with linear regression ($r^2$) and logistic regression (accuracy). When OOD dataset is separated and used as test dataset in  $(^b)$ the performance of the model decreases. 
% OOD case in the epsilon $(^*)$ dataset cannot be identified. 
}
\begin{tabular}{l|c|c|r|rr|rr} \hline
\multirow{2}{*}{Dataset} & \multirow{2}{*}{Det} & \multirow{2}{*}{Norms} & \multirow{2}{*}{Threshold} & \multicolumn{2}{l}{ID Accuracy} & \multicolumn{2}{l}{OOD Accuracy} \\
                         &                      &                        &                            & Train          & Test           & Train          & Test            \\ \hline
Adult                    & O                    & L2                     & 0.1628                     & 0.7825         & 0.7830          & 0.7978         & 0.2674          \\
Helena                   & O                    & L1                     & 0.045                      & 0.1945         & 0.1966         & 0.1464         & 0.0475          \\
Jannis                   & T                    & L1                     & -0.02                      & 0.5619         & 0.5639         & 0.5772         & 0.4742          \\
Higgs small              & O                    & L1                     & 0.042                      & 0.6222         & 0.6168         & 0.6226         & 0.5684          \\
Aloi                     & O                    & L1                     & 0.016                      & 0.2606         & 0.2325         & 0.3282         & 0.0711          \\
Covtype                  & T                    & L1                     & -0.035                     & 0.6046         & 0.6043         & 0.6035         & 0.4354          \\
California               & O                    & L1                     & 0.11                       & 0.3334         & 0.1806         & 0.4771         & -6.0335         \\
Year                     & O                    & L2                     & 0.045                      & 0.1686         & 0.1670          & 0.1697         & -0.7141         \\
Yahoo                    & T                    & L1                     & 1.46eE-03                  & 0.3256         & 0.3262         & 0.3255         & -0.2978         \\
Microsoft                & T                    & L1                     & -8.30eE-03                 & 0.0456         & 0.0441         & 0.0473         & -0.8418    \\ \hline  
\end{tabular}
\label{tab:oodSetting}
\end{table*}

Table  \ref{tab:oodSetting}  shows significant differences between the two settings. Without OOD (Table  \ref{tab:oodSetting}, Section a), the training and test results are comparable. However, when used as test data, the OOD reduces the performance of the models. OOD leads to a 20\% decrease (Table  \ref{tab:oodSetting}, Section b) in performance between training and test results for the classification task, and a negative r2 for the regression task.

\subsection{Models Performance}

Table \ref{tab:results} shows the results of the experiment. 
\begin{table*}
% \small
    \centering
    \caption{Experiment result. F1 score for classification and RMSE for regression. Datasets with (*) mean a regression problem. Models with ($^c$) are contrastive learning based models.}
    \begin{tabular}{l|cccccc|cccc} \hline 
         &  AD$\uparrow$&  HE$\uparrow$&  JA$\uparrow$&  HI$\uparrow$&  AL$\uparrow$& CO$\uparrow$&   CA*$\downarrow$&YE*$\downarrow$&YA*$\downarrow$&MI*$\downarrow$\\ \hline  
         FT-T&  0.782&  0.153&  0.572&  0.738&  0.407&  -&   0.867&\textbf{6.461}&-&-\\ 
         DCN2&  0.744&  0.129&  0.542&  0.710&  0.414&  0.58&   2.602&7.054&0.645&0.746\\   
 GrowNet& 0.465& -& -& 0.685& -& -& 0.969& 7.605& 1.01&0.769\\  
 ResNet& 0.652& 0.10& 0.574& 0.753& 0.437& 0.694& 0.892& 6.496& \textbf{0.639}&\textbf{0.736}\\ 
 MLP& 0.508& 0.146& 0.561& 0.753& 0.326& 0.617& 0.894& 6.488& 0.657&0.741\\  
 AutoInt& 0.78& 0.133& 0.549& 0.719& 0.401& 0.608& 0.89& 6.673& -&0.739\\
 TabR-MLP& 0.688& 0.165& 0.541& 0.753& 0.429& 0.688& 2.677& 2e5& 1.285&0.79\\ \hline
 TCL$^c$& \textbf{0.831}& \textbf{0.154}& \textbf{0.575}& \textbf{0.758}& \textbf{0.447}& \textbf{0.880}& \textbf{0.843}&6.491&0.652&0.738\\
 Scarf$^c$& 0.720& 0.00& 0.122& 0.308& 0.00& 0.091& -& -& -&-\\
 SubTab$^c$& 0.714& 0.146& 0.504& 0.602& 0.322& 0.59&   1.012&6.668&0.656&0.744\\ \hline
    \end{tabular}
    \label{tab:results}
\end{table*}

Overall, TCL outperforms other models, while the performance of the other models is comparable across various datasets. There are some exceptions where specific models underperform relative to others. For instance, GrowNet performs below average on the adult dataset, DCN V2 underperforms on the California Housing dataset, and GrowNet also underdelivers on the Yahoo dataset. In contrast, TCL stands out by outperforming other models on most datasets, particularly in classification problems. Nevertheless, TCL's performance in regression problems is not significantly behind that of the top models.

\subsection{Training Duration}
Table  \ref{tab:resultsTime} displays the training duration for the best three deep learning models. Each model has unique characteristics and training steps, and all seven models (FT-T, DCN 2, GrowNet, ResNet, MLP, AutoInt, TabR-MLP) underwent extensive tuning. The Yahoo and Microsoft datasets required 5 days to complete the entire parameter-tuning process. For FTT and restnet, a single training time was sampled once the tuning process was completed. TCL, which involves unsupervised training, the time recorded is the time for each model to stabilise their loss with a 256 batch size, which is around 15 epochs. it is clear that TCL has a short training time.
\begin{table*}
% \small
    \centering
    \caption{Table of training duration in seconds of each dataset. Datasets with (*) mean a regression problem. All models except for TCL are trained with a GPU. TCL were trained with CPU}
    \begin{tabular}{l|cccccc|cccc} \hline 
         &  AD$\downarrow$&  HE$\downarrow$&  JA$\downarrow$&  HI$\downarrow$&  AL$\downarrow$& CO$\downarrow$&   CA*$\downarrow$&YE*$\downarrow$&YA*$\downarrow$&MI*$\downarrow$\\ \hline  
         FT-T&  1027&  130&  155&  94&  1205&  -&   88&1290&-&-\\ 
 ResNet& 2.1e+02& 32& 21& 56& 44& 618& 15& 236& 284&950\\ 
 TCL& 15& 23& 23& 38& 40& 330&   7&240&620&820\\ \hline
    \end{tabular}
    
    \label{tab:resultsTime}
\end{table*}
\subsection{Efficiency Evaluations}
\begin{table*}[]
% \small
\centering
\caption{A speed/accuracy trade off matrix $T = \frac{P}{t}$ where $P$ performance matrix used and $t$ is time in second required.  A higher result is better. Datasets with (*) mean a regression problem.}
\begin{tabular}{r|rrrrrr|rrrr}
\toprule
 & AD      & HE     & JA     & HI     & AL      & CO     & CA*     & YE*     & YA*     & MI*     \\
\midrule
FT-T  & 0.00076 & 0.0012 & 0.0037 & 0.0079 & 0.00034 & -      & 0.013& 0.00012& -       & -       \\
ResNet& 0.0031& 0.0031& 0.027& 0.013& 0.0099& 0.0011& 0.075& 0.00065& 0.0055& 0.0014\\
TCL& 0.055& 0.0066& 0.025& 0.028& 0.0199& 0.0026& 0.16& 0.00064& 0.0024& 0.0016\\ \hline

\end{tabular}

\label{tab:tradeOff}
\end{table*}
Table \ref{tab:tradeOff} shows that TCLs are dominant. FT-Transformer and ResNet produce a good F1 and RMSE score; however, they take more time to train. 
In \textbf{FT-Transformer}, multiple attention heads process numeric and categorical features separately before combining them. The model includes four types of layers that grow exponentially, leading to resource-intensive computations.
\textbf{ResNet} employs parallel calculations across multiple convolutional layers (SubNet), using three identical SubNets, one of which is highly filtered. In contrast, TCL has a simpler architecture akin to MLP, achieving a high speed/accuracy trade-off.
\textbf{TCL} features narrow layers for both the encoder and decoder, each with one hidden layer and one normalisation layer, resulting in fewer layers than ResNet. However, TCL's pair operation for loss calculation doubles its training time.

\section{Conclusion }
The choice of models for tabular datasets with out-of-distribution (OOD) data depends on the user's needs and available resources. TCL outperforms other heavier models for classification problems on OOD while maintaining efficiency. RestNet and FT-Transformer perform well on many datasets, but these models require more resources, which may not always be feasible. It is worth noting that TCL was trained on a CPU, and RestNet and FT-T were trained on a GPU.  This makes TCL available for more users than other models that require more training resources. Both RestNet and TCL can be options for fine-tuning and serving as head-to-head comparison models. 

Although TCL has shown promising results, there are opportunities for potential enhancement. A continual learning can be proposed to improve performance.  Further optimisation of the contrastive learning process can be studied to achieve even greater efficiency. Additionally, there is a need to explore TCL's performance on a wider range of domain-specific tabular datasets. Furthermore, it is crucial to investigate TCL's interpretability, as this is important for many real-world applications. Finally, the code for our experiment can be found online \cite{tclCode}.

% \section{Reproducibility}
% The code for this work can be found online \cite{tclCode} (submitted as a supplementary file). The dataset is also available online and can be downloaded using the information provided in the citation.

\section{Acknowledgements}
This work acknowledges the funding provided by the Indonesia Endowment Funds for Education (LPDP) and the support from the Indonesian Taxation Office and the University of Queensland, Australia.

\bibliographystyle{ieeetr}
\bibliography{sample}

\begin{thebibliography}{10}

\bibitem{Hsu2020}
Y.-C. Hsu, Y.~Shen, H.~Jin, and Z.~Kira, ``Generalized odin: Detecting out-of-distribution image without learning from out-of-distribution data,'' 2020.

\bibitem{OODBaseline}
D.~Hendrycks and K.~Gimpel, ``A baseline for detecting misclassified and out-of-distribution examples in neural networks,'' {\em 5th International Conference on Learning Representations, ICLR 2017 - Conference Track Proceedings}, 10 2016.

\bibitem{Hwang2018Background}
T.~Hwang, ``Computational power and the social impact of artificial intelligence,'' {\em SSRN Electronic Journal}, 2018.

\bibitem{ahmed2020dedemocratizationaideeplearning}
N.~Ahmed and M.~Wahed, ``The de-democratization of ai: Deep learning and the compute divide in artificial intelligence research,'' 2020.

\bibitem{Lee2020}
D.~Lee, S.~Yu, and H.~Yu, ``{Multi-Class Data Description for Out-of-distribution Detection},'' {\em Proceedings of the ACM SIGKDD International Conference on Knowledge Discovery and Data Mining}, pp.~1362--1370, aug 2020.

\bibitem{multiClass}
D.~Lee, S.~Yu, and H.~Yu, ``Multi-class data description for out-of-distribution detection,'' in {\em Proceedings of the 26th ACM SIGKDD International Conference on Knowledge Discovery \& Data Mining}, KDD '20, (New York, NY, USA), p.~1362–1370, Association for Computing Machinery, 2020.

\bibitem{Bendale2016}
A.~Bendale and T.~E. Boult, ``Towards open set deep networks,'' {\em Proceedings of the IEEE Computer Society Conference on Computer Vision and Pattern Recognition}, vol.~2016-December, pp.~1563--1572, 12 2016.

\bibitem{Gal2016}
Y.~Gal and Z.~Ghahramani, ``Dropout as a bayesian approximation: Representing model uncertainty in deep learning,'' 6 2016.

\bibitem{Liang2017}
S.~Liang, Y.~Li, and R.~Srikant, ``Enhancing the reliability of out-of-distribution image detection in neural networks,'' {\em 6th International Conference on Learning Representations, ICLR 2018 - Conference Track Proceedings}, 6 2017.

\bibitem{Grinsztajn2022}
L.~Grinsztajn, E.~Oyallon, and G.~Varoquaux, ``{Why do tree-based models still outperform deep learning on typical tabular data?},'' {\em Advances in Neural Information Processing Systems}, vol.~35, pp.~507--520, dec 2022.

\bibitem{anonymizedMethod}
A.~Ginanjar, X.~Li, and W.~Hua, ``Contrastive federated learning with tabular data silos,'' 2024.

\bibitem{Chen2020SimClr}
T.~Chen, S.~Kornblith, M.~Norouzi, and G.~Hinton, ``A simple framework for contrastive learning of visual representations,'' pp.~1597--1607, 11 2020.

\bibitem{subtab}
T.~Ucar, E.~Hajiramezanali, and L.~Edwards, ``Subtab: Subsetting features of tabular data for self-supervised representation learning,'' vol.~23, 2021.

\bibitem{huang2017speedaccuracy}
J.~Huang, V.~Rathod, C.~Sun, M.~Zhu, A.~Korattikara, A.~Fathi, I.~Fischer, Z.~Wojna, Y.~Song, S.~Guadarrama, and K.~Murphy, ``Speed/accuracy trade-offs for modern convolutional object detectors,'' 2017.

\bibitem{platt1999probabilistic}
J.~Platt {\em et~al.}, ``Probabilistic outputs for support vector machines and comparisons to regularized likelihood methods,'' {\em Advances in large margin classifiers}, vol.~10, no.~3, pp.~61--74, 1999.

\bibitem{Guo2017}
C.~Guo, G.~Pleiss, Y.~Sun, and K.~Q. Weinberger, ``On calibration of modern neural networks,'' {\em 34th International Conference on Machine Learning, ICML 2017}, vol.~3, pp.~2130--2143, 6 2017.

\bibitem{kirchheim2022pytorch}
K.~Kirchheim, M.~Filax, and F.~Ortmeier, ``Pytorch-ood: A library for out-of-distribution detection based on pytorch,'' in {\em Proceedings of the IEEE/CVF Conference on Computer Vision and Pattern Recognition (CVPR) Workshops}, pp.~4351--4360, June 2022.

\bibitem{deepMCCD}
D.~Lee, S.~Yu, and H.~Yu, ``Multi-class data description for out-of-distribution detection,'' 2020.

\bibitem{mlp}
D.~W. Ruck, S.~K. Rogers, and M.~Kabrisky, ``Feature selection using a multilayer perceptron,'' {\em Journal of neural network computing}, vol.~2, no.~2, pp.~40--48, 1990.

\bibitem{tabr}
Y.~Gorishniy, I.~Rubachev, N.~Kartashev, D.~Shlenskii, A.~Kotelnikov, and A.~Babenko, ``Tabr: Tabular deep learning meets nearest neighbors in 2023,'' 2023.

\bibitem{snn}
G.~Klambauer, T.~Unterthiner, A.~Mayr, and S.~Hochreiter, ``Self-normalizing neural networks,'' {\em 31st Conference on Neural Information Processing Systems}, 2017.

\bibitem{vaswani2017attentiontransformer}
A.~Vaswani, N.~Shazeer, N.~Parmar, J.~Uszkoreit, L.~Jones, A.~N. Gomez, {\L}.~Kaiser, and I.~Polosukhin, ``Attention is all you need,'' {\em Advances in neural information processing systems}, vol.~30, 2017.

\bibitem{restnet}
B.~Li, W.~Wei, A.~Ferreira, and S.~Tan, ``Rest-net: Diverse activation modules and parallel subnets-based cnn for spatial image steganalysis,'' {\em IEEE Signal Processing Letters}, vol.~25, no.~5, pp.~650--654, 2018.

\bibitem{DCNV2}
R.~Wang, R.~Shivanna, D.~Z. Cheng, S.~Jain, D.~Lin, L.~Hong, and E.~H. Chi, ``Dcn v2: Improved deep and cross network and practical lessons for web-scale learning to rank systems,'' {\em The Web Conference 2021 - Proceedings of the World Wide Web Conference, WWW 2021}, pp.~1785--1797, 8 2020.

\bibitem{AutoInt}
W.~Song, C.~Shi, Z.~Xiao, Z.~Duan, Y.~Xu, M.~Zhang, and J.~Tang, ``Autoint: Automatic feature interaction learning via self-attentive neural networks,'' {\em International Conference on Information and Knowledge Management, Proceedings}, vol.~10, pp.~1161--1170, 10 2018.

\bibitem{NODE}
S.~Popov, S.~Morozov, and A.~Babenko, ``Neural oblivious decision ensembles for deep learning on tabular data,'' {\em 8th International Conference on Learning Representations, ICLR 2020}, 9 2019.

\bibitem{TabNet}
S.~Arık and T.~Pfister, ``Tabnet: Attentive interpretable tabular learning,'' {\em 35th AAAI Conference on Artificial Intelligence, AAAI 2021}, vol.~8A, pp.~6679--6687, 8 2019.

\bibitem{GrowNet}
S.~Badirli, X.~Liu, Z.~Xing, A.~Bhowmik, K.~Doan, and S.~S. Keerthi, ``Gradient boosting neural networks: Grownet,'' 2 2020.

\bibitem{scarf}
D.~Bahri, H.~Jiang, Y.~Tay, and D.~Metzler, ``{SCARF: SELF-SUPERVISED CONTRASTIVE LEARNING USING RANDOM FEATURE CORRUPTION},'' in {\em ICLR 2022 - 10th International Conference on Learning Representations}, 2022.

\bibitem{adult_2}
B.~Becker and R.~Kohavi, ``{Adult}.'' UCI Machine Learning Repository, 1996.
\newblock {DOI}: https://doi.org/10.24432/C5XW20.

\bibitem{helenaJannis}
I.~Guyon, L.~Sun-Hosoya, M.~Boull{\'e}, H.~J. Escalante, S.~Escalera, Z.~Liu, D.~Jajetic, B.~Ray, M.~Saeed, M.~Sebag, A.~Statnikov, W.-W. Tu, and E.~Viegas, ``Analysis of the automl challenge series 2015-2018,'' in {\em AutoML}, Challenges in Machine Learning, Springer, 2019.

\bibitem{higgssmall}
P.~Baldi, P.~Sadowski, and D.~Whiteson, ``Searching for exotic particles in high-energy physics with deep learning,'' {\em Nature Communications}, vol.~5, p.~4308, 2014.

\bibitem{aloi}
J.-M. Geusebroek, G.~J. Burghouts, and A.~W.~M. Smeulders, ``The amsterdam library of object images,'' {\em International Journal of Computer Vision}, vol.~61, no.~1, pp.~103--112, 2005.

\bibitem{covtype}
J.~A. Blackard and D.~J. Dean, ``Comparative accuracies of artificial neural networks and discriminant analysis in predicting forest cover types from cartographic variables,'' {\em Computers and Electronics in Agriculture}, vol.~24, no.~3, pp.~131--151, 2000.

\bibitem{ca}
R.~K. Pace and R.~Barry, ``Sparse spatial autoregressions,'' {\em Statistics \& Probability Letters}, vol.~33, no.~3, pp.~291--297, 1997.

\bibitem{year}
T.~Bertin-Mahieux, D.~P. Ellis, B.~Whitman, and P.~Lamere, ``The million song dataset,'' in {\em Proceedings of the 12th International Conference on Music Information Retrieval (ISMIR 2011)}, (Miami, Florida, USA), pp.~591--596, October 2011.

\bibitem{chapelle2011yahoo}
O.~Chapelle and Y.~Chang, ``Yahoo! learning to rank challenge overview,'' in {\em Proceedings of the Learning to Rank Challenge}, vol.~14 of {\em Proceedings of Machine Learning Research}, pp.~1--24, PMLR, 2011.

\bibitem{microsoft}
T.~Qin and T.-Y. Liu, ``Introducing {LETOR} 4.0 datasets,'' {\em arXiv preprint arXiv:1306.2597}, 2013.

\bibitem{tclCode}
A.~Ginanjar, ``Tabular contrastive learning (tcl).'' [Online]. Available from: \url{https://github.com/mambo06/TCL}, July~12 2024.

\end{thebibliography}
\end{document}